\documentclass{article} % For LaTeX2e
\usepackage[final]{colm2026_conference}
\usepackage{fontawesome5}

\usepackage{microtype}
\usepackage{hyperref}% load hyperref before cleveref
\usepackage{url}
\usepackage{tabularx}
\usepackage{graphicx}
\usepackage{subcaption}
\usepackage{booktabs} % for professional tables
\usepackage{siunitx}
\usepackage{multirow}
\usepackage{enumitem}
\usepackage{bm} % for bold math symbols
\usepackage{float}
\usepackage{tcolorbox}
\usepackage{amsmath}
\usepackage[capitalise,noabbrev]{cleveref}

\usepackage{lineno}
\usepackage{xcolor}

\definecolor{darkblue}{rgb}{0, 0, 0.5}
\hypersetup{colorlinks=true, citecolor=darkblue, linkcolor=darkblue, urlcolor=darkblue}

\title{LLMs Encode Their Failures: Predicting Success \\ from Pre-Generation Activations}

\author{ William Lugoloobi$^{1,}$\thanks{Corresponding author: \texttt{william.lugoloobi@oii.ox.ac.uk}}, Thomas Foster$^{2}$\thanks{equal contribution}, William Bankes$^{3}$\footnotemark[2], Chris Russell$^{1}$ \\ \\ $^{1}$ Oxford Internet Institute, University of Oxford \\ $^{2}$ FLAIR, University of Oxford \\ $^{3}$ Department of Computer Science, University College London \\}

\newcommand{\hficon}{%
  \raisebox{-0.18em}{%
    \includegraphics[
      height=1.15em,
      trim=15 15 15 15,
      clip
    ]{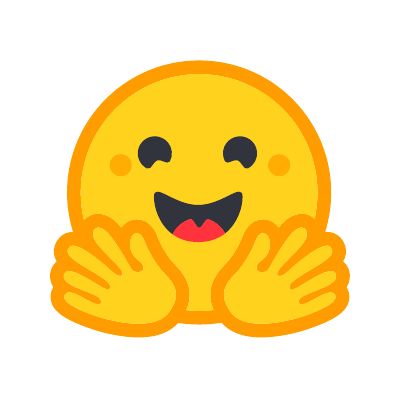}%
  }%
}

\begin{document}

% \ifcolmsubmission
\ifcolmfinal
% \linenumbers
\fi

\maketitle

% \vspace{-0.6em}
\begin{center}
\href{https://github.com/KabakaWilliam/llms_know_difficulty}
{{\color{black}\faGithub}\ \textbf{Code}}
\hspace{1.8em}
\href{https://huggingface.co/CoffeeGitta/pika-probes}
{\hficon\ \textbf{Probes}}
\end{center}

% \vspace{-0.5em}

\begin{abstract}
Running LLMs with extended reasoning on every problem is expensive, but determining which inputs actually require additional compute remains challenging. We investigate whether a model's likelihood of success is recoverable from its internal representations prior to generation, and whether this signal can guide more efficient inference. We train linear probes on pre-generation activations to predict policy-specific success on math and coding tasks, substantially outperforming surface features such as question length and TF-IDF. Using E2H-AMC, which provides both human and model performance on identical problems, we show that models encode a model-specific notion of difficulty that is distinct from human difficulty, and that this distinction increases with extended reasoning. Leveraging these probes, we demonstrate that routing queries across a pool of models can match the performance of the best-performing model whilst reducing inference cost by up to 70\% on MATH, showing that internal representations enable practical efficiency gains even when they diverge from human intuitions about difficulty.
% at \url{https://anonymous.4open.science/r/anonmyous_difficulty-D5EA}.

\end{abstract}

\section{Introduction}
\label{sec:introduction}

Large Language Models (LLMs) have achieved remarkable performance on mathematics and programming tasks~\citep{hendrycksmath2021, cobbe_training_2021, naman_jain_livecodebench_2024}. %, where correctness can be objectively verified.
Because model outputs are typically generated via stochastic decoding, performance is naturally characterized by a \emph{success rate}---the probability that a model will correctly answer a given query.
Accurately estimating success rates is critical for model routing systems that direct queries to the model most likely to succeed~\citep{chen_frugalgpt_2024, ding_hybrid_2023}, but obtaining low-variance estimates requires multiple costly rollouts per input.
This invites the question: Can we predict whether a model will succeed \emph{before} it generates any output?

 We show that LLMs internally encode estimates of their own success in pre-generation activations, and that these estimates can be efficiently extracted using linear probes.
Prior work has shown that models contain correctness-related signals~\citep{kadavath_language_2022, azaria_internal_2023, burns_discovering_2024}, but it remains unclear what notion of difficulty these signals represent and whether they are reliable enough for practical decision-making.

We conduct an empirical study across mathematics (MATH, GSM8K, AIME, E2H-AMC) and coding (LiveCodeBench) domains, training linear probes on pre-generation activations to predict success under various decoding policies.
Our investigation reveals that LLMs encode a \emph{model-specific} notion of difficulty that differs systematically from human judgments and varies with the inference-time policy.
We demonstrate that probe-guided routing can match high-compute accuracy at 70\% cost reduction, while identifying critical failure modes where probe reliability becomes the bottleneck.

\paragraph{Main Contributions.}
\begin{itemize}[leftmargin=*, itemsep=3pt]
    \item \textbf{Human and model difficulty are encoded differently in LLMs.}
    Using the AMC subset of Easy2Hard-Bench (E2H-AMC), which provides both human difficulty labels (estimated via Item Response Theory, IRT) and model accuracy on identical questions, we show that linear probes can extract both signals from pre-generation activations (Spearman $\rho = 0.83$--$0.87$ for human difficulty, $0.40$--$0.64$ for model difficulty).
    Crucially, these represent \emph{distinct information}: model-derived difficulty proves more predictive of actual performance, and as models solve harder problems through extended reasoning, their internal representations increasingly diverge from human difficulty judgments (Section~\ref{sec:predicting_difficulty}).
    
    \item \textbf{Probes reliably predict model performance across decoding settings and reasoning modes.}
    Binary classification of success under fixed decoding policies (greedy, Maj@K) achieves strong discrimination (AUROC $> 0.7$ for several models) and remains stable across sampling temperatures and majority voting thresholds.
    However, probe reliability degrades with extended test-time compute, though this is partially recoverable with non-linear MLP probes (Section~\ref{sec:predicting_difficulty}).
    
    \item \textbf{Probe-guided routing achieves substantial cost savings with minimal accuracy loss.}
    Simple threshold-based and utility-maximizing routing policies match the highest-capability single-model performance at 70\% lower inference cost on MATH, with similar gains on AIME and GSM8K.
    In some configurations, our router exceeds the best baseline while approaching oracle-level accuracy, demonstrating that reliable difficulty estimates—not routing sophistication—are the key to effective model allocation (Section~\ref{sec:routing}).
\end{itemize}

\section{Related Work}

\textbf{Predicting model correctness.} For routing, abstention, and compute allocation decisions, we need estimates of whether a model will answer correctly. Prior work shows that LLMs contain correctness-related signals. \cite{kadavath_language_2022} demonstrate that models can predict their own correctness when explicitly prompted for P(True)-style self-assessments, but this requires generation overhead unsuitable for routing. A complementary line identifies "truthfulness" or "correctness directions" in internal activations \citep{azaria_internal_2023, burns_discovering_2024, li_inference-time_2023}, which implicitly estimate confidence by predicting correct responses \citep{geng_survey_2024}. Most directly related, \citet{cencerrado_no_2025} extract correctness directions via difference-of-means between pass and fail activation centroids, then test whether these directions transfer to predict success on new questions. They find strong performance in factual settings but substantially weaker results on mathematical reasoning (GSM8K AUROC $\approx$ 0.6–0.7).
We take a different approach: rather than extracting unsupervised directions, we train supervised linear classifiers directly on labeled pass/fail examples to predict binary success. This supervised formulation achieves stronger discrimination on reasoning tasks (AUROC $>$ 0.7 for several models) and enables us to systematically investigate: (1) what notion of difficulty these probes encode, and (2) how probe reliability varies with extended reasoning.

\textbf{Difficulty estimation.} Recent work shows that pre-generation activations contain linearly decodable difficulty signals \citep{lugoloobi_llms_2025, lee_probing_2025}, but it remains unclear whether these represent human difficulty, model-specific difficulty, or both. \citet{lugoloobi_llms_2025} demonstrate that models encode problem difficulty but focus primarily on correlation with IRT \citep{embretson2000item, woodruff_estimation_1996} scores—psychometric measures calibrated from large-scale human performance data.\citet{lee_probing_2025} probe difficulty perception mechanisms without systematically comparing human versus model difficulty or evaluating routing applications. We provide the first direct comparison using the AMC subset of Easy2Hard-Bench \citep{ding_easy2hard-bench_2024}, where human IRT scores and model performance are available, %on identical questions, 
establishing that these are \emph{distinct} signals. Critically, we show this divergence intensifies with extended reasoning—models allocate computation according to human difficulty even when reliably solving those problems.

\textbf{Test-time compute scaling.} Sampling-based methods like self-consistency \citep{wang_self-consistency_2022} aggregate k reasoning paths through majority voting (maj@k) to improve accuracy on complex reasoning tasks. \citet{cobbe_training_2021} train verifiers to re-rank generated solutions, substantially improving performance on math tasks. Recent models with extended reasoning capabilities (e.g., DeepSeek-R1, o1-series) scale test-time compute by generating longer chain-of-thought responses \cite{guo2025deepseek, openai_learning_2024}. While prior work focuses on the accuracy-compute tradeoff, we provide the first systematic investigation of how test-time scaling—both through majority voting and extended reasoning—affects the linear accessibility of difficulty information in pre-generation activations. Our finding that probe quality degrades with increased reasoning budget (AUROC: 0.78 → 0.64) despite improved accuracy (86.6\% → 92.0\%) has implications for adaptive inference systems that rely on difficulty estimates extracted before generation.

\textbf{Model routing.} Prior routing work relies on indirect proxies such as input length, perplexity, or heuristic confidence measures \citep{chen_frugalgpt_2024, ding_hybrid_2023}. \citet{chen_frugalgpt_2024} use multiple API calls to estimate confidence, while \citet{ding_hybrid_2023} route based on input complexity heuristics. \citet{song_irt-router_2025} propose an IRT-based approach that models question difficulty and model ability via learned latent traits from embeddings. In contrast, our probe-based method directly predicts model-specific success from pre-generation representations, requiring no additional generation or separate embedding models at routing time. This yields 17–70\% cost savings while matching high-capability model performance. Critically, we find that routing effectiveness is limited by the reliability of the underlying success estimates, not the routing policy itself.
\section{Predicting Difficulty}
\label{sec:predicting_difficulty}

% \will{Most comments start here, will go back and look at the Intro and Related Work when time. The paragraph below feels like it belongs in the Introduction...}

% Adaptive systems for model routing and training-data selection depend on accurate difficulty prediction.
% Prior work shows that pre-generation activations contain linearly decodable signals that anticipate downstream performance and correlate with perceived difficulty~\cite{lugoloobi_llms_2025, cencerrado_no_2025, lee_probing_2025}.
% However, it remains unclear what this signal represents: human difficulty, model-specific difficulty under a particular decoding policy, or a conflation of both.
% %
% In this section we disentangle these notions.
% Using E2H-AMC from the Easy2HardBench dataset \citep{ding_easy2hard-bench_2024}, where we have human IRT difficulty labels and can also estimate model success on the \emph{same} questions via rollouts, we train linear probes for each target from identical pre-generation activations and show that they are not the same signal.

% Difficulty of an environment is dependent on the ability of an agent operating within that environment. 

% difficulty can be how frequently a problem is solved or if given many tires an agent gets any solution

Difficulty is subjective: it depends on the actor facing the task. Just as students in a class find different questions hard, models have their own strengths and weaknesses. Difficulty also depends on what we count as success; a task that is hard on a single attempt may be much easier when several are allowed. Given this subjectivity, we consider a range of human- and model-derived difficulty definitions, and introduce activation probes as a tool for examining how a model perceives each of them.

\subsection{Different Notions of Difficulty}
\label{sec:two_difficulties}

\paragraph{Human difficulty.}
IRT models question difficulty in a similar manner to Elo in games such as chess. The IRT difficulty score $b(q) \in [0,1]$ ranks questions from easiest (0) to hardest (1).  
% \paragraph{Human difficulty (IRT).} \will{Should we introduce the concept of difficulty independent from a specific dataset?}
% On E2H-AMC, each question $q$ is annotated with a human IRT difficulty $b(q)$, where larger values indicate questions that are harder for humans. \will{Do we have any notion of which humans the difficulty represents? - student performance}

\paragraph{Model difficulty: Expected success rate.}
Another notion of difficulty is that of the expected success rate, which captures the probability of a stochastic agent succeeding at a task or challenge. For a model with a stochastic decoding policy $\pi$ and question $q$ with a ground-truth answer $y^*$, we define the expected success rate as
\begin{align}
    s(\pi, q) \;=\; \mathbb{E}_{a \sim \pi(\cdot \mid q)}\big[\mathbb{I}(\mathrm{parser}(a) = y^*)\big],
\end{align}
where $\mathrm{parser}(\cdot)$ extracts a final answer from response $a$.
We estimate $s(\pi,q)$ with $K$ Monte Carlo rollouts:
\begin{align}
\hat{s}_{\mathrm{MC}}(\pi, q)
\;=\;
\frac{1}{K}\sum_{k=1}^{K}
\mathbb{I}\big(\mathrm{parser}(a_k) = y^*\big),
\end{align}
%with $a_k \sim \pi(\cdot \mid q)$ i.i.d. \will{Here we include experiment settings with theoretical concepts, separate the two, introduce the concept then introduce the experiment setup + models etc...} We use $T=1$ and $K=50$ ($K=5$ for GPT-OSS due to compute cost during generation). This formulation provides a continuous measure of model-specific difficulty that ranks questions by expected performance under stochastic decoding.
with $a_k \sim \pi(\cdot \mid q)$ i.i.d. This formulation provides a continuous measure of model-specific difficulty that ranks questions by expected performance under stochastic decoding.

\paragraph{Model success: Binary outcome under specified decoding.} Difficulty can also be thought of from a binary perspective; thus we also consider binary success under different deterministic aggregation rules commonly used by LLMs. 
%\will{Maybe don't tie this to routing etc... that's an application this is a theoretical notion of difficulty} For routing and other decision-making applications, we also consider binary success under deterministic aggregation rules.
Specifically, we evaluate:
\begin{itemize}[leftmargin=*, itemsep=2pt]
    \item \textbf{Greedy}: The model either succeeds or fails on a single deterministic generation.
    \item \textbf{Maj@K} (Majority voting): Select the most frequent parsed answer from $K$ samples and verify correctness.
    \item \textbf{Pass@K}: The question is considered answered if at least one of the $K$ sampled answers is correct \citep{chen2021evaluating}.  
\end{itemize}

% not quite true both produce probabilities but from different amounts of information...

%Unlike $\hat{s}_{\mathrm{MC}}$, which estimates a probability, these targets predict whether a specific inference procedure will succeed. Because Maj@K depends on the full answer distribution rather than individual samples, it captures different information about model capability and serves as a distinct prediction target.

Whilst both the binary and expected success metrics capture the probability of a model being successful, they are different transformations of the full answer distribution and thus capture different information about model capacity, serving as distinct prediction targets.

%\will{Do we go to experimental setup from here? We haven't introduced the probes, our learning objective for the different difficulty notions or established what we're doing experiments on...}

\subsection{Activation Probes}
\label{sec:activation probes}

\begin{center}
\vspace{-0.1cm}
\begin{tcolorbox}[colback=gray!10, colframe=gray!50, boxrule=0.5pt,
                  boxsep=2pt, top=3pt, bottom=3pt]
\small
\textbf{Example suffixes:}\\[3pt]
\textbf{Qwen2.5:} \texttt{<|im\_end|>\textbackslash n<|im\_start|>assistant\textbackslash n} \\
\textbf{DeepSeek-R1:} \texttt{<|Assistant|><think>\textbackslash n} \\
\textbf{GPT-OSS:} \texttt{<|end|><|start|>assistant}
\end{tcolorbox}
\captionof{figure}{Example chat-template suffixes for several model families.}
\label{fig:probe suffix positions}
\vspace{-0.1cm}
\end{center}

% How we extract activations. Define the probe and activations

Let $h_i^{\ell} \in \mathbb{R}^D$ be the residual-stream activation, where $D$ is the size of the hidden dimension, $\ell$ the layer index of the model, and $i \in [S]$ the sequence position of a prompt with length $S$. Depending on the notion of difficulty we are probing we use different linear probe setups. For expected success rate and human IRT difficulty we use the MSE loss and for binary success under different decoding strategies we use a logistic regression probe with a cross-entropy loss. Across all objectives we include an L2-regularization term in the loss. 

During training we perform 5-fold cross validation on the probe, and sweep three hyperparameters: the layer index $\ell$, the L2 regularisation weight $\alpha$, and the token position $i$, selected from the final P non-padding tokens of the chat template \citep{arditi_refusal_2024}; see \Cref{fig:probe suffix positions}.

% What objectives we use for training i.e. regression, cross-entropy etc...

% How we define the label and input.

% Training procedure etc... K-fold CV

%We investigate how models represent difficulty with activation probes. Let $A \in \mathbb{R}^{S \times D}$ denote residual stream activations (pre-layer norm) from a fixed layer $l$, where $S$ is the sequence length of an input, and $D$ the hidden dimension of the model. Following \citet{arditi_refusal_2024}, we extract activations at post-instruction template positions i.e. the final tokens before generation begins but after the user input (see Appendix \will{reference existing appendix}) to get a vector $a \in \mathbb{R}^D \subset A$. \will{check my math notation...}  

%We train simple linear probes from each layer and position using an 80/20 train–validation split for hyperparameter selection, and report our best probe results on a held-out test set. For success-rate prediction, we use MSE loss; for binary success (Maj@K, greedy) we use binary cross-entropy. We apply Platt scaling on a validation set to calibrate probabilities from our best-trained classification probes. Additional training details are in Appendix~\ref{app:Probing_Formulation}

\section{Experiments}

\subsection{Experimental Setup}
\label{sec:difficulty_setup}

\textbf{Models.} We evaluate on a heterogeneous pool of models spanning sizes and reasoning capabilities: GPT-OSS-20B (low/medium/high reasoning)~\citep{openai2025gptoss}, DeepSeek-R1-Distill-Qwen-7B~\citep{guo2025deepseek}, Qwen2.5-Math (1.5B, 7B) \citep{yang2024qwen25math}, Qwen2.5-Coder (3B, 7B) \citep{hui_qwen25-coder_2024}, and Qwen2.5-1.5B \citep{team_qwen25_2024}. Full generation configurations are provided in Appendix~\ref{sec:model_settings}.

\textbf{Datasets.} We construct a range of datasets to train model-difficulty probes. For E2H-AMC, GSM8K, MATH, AIME (1983-2024), and LiveCodeBench~\citep{ding_easy2hard-bench_2024, cobbe_training_2021, hendrycksmath2021, veeraboina_gneubigaime-1983-2024_2023, balunovic_matharena_2025, naman_jain_livecodebench_2024}; we generate multiple answers for each of the models above. Using these answers we construct a label for each prompt, as defined in \Cref{sec:two_difficulties}. Given the quantity of data and expense of generating multiple rollouts for the larger GPT-OSS model, we set K=5 (the number of Monte Carlo rollouts per prompt) for this model, and K=50 for the smaller Qwen based models. For LiveCodeBench we use temporal splits based on each model's release date to control for data contamination in our analysis. Where test and validation splits are available we use those provided, otherwise we split the data into test splits using an 80/20 ratio. AIME-2025 \citep{balunovic_matharena_2025} is used only as a test set.  Full dataset details can be found in Appendix~\ref{app:dataset statistics}.

A particularly important dataset in our analysis is \emph{the E2H-AMC dataset} \cite{ding_easy2hard-bench_2024}, which contains 4k mathematics problems annotated with an IRT difficulty score $b(q)$, calibrated from large-scale student performance data from secondary school students who sat the AMC. This provides a model-agnostic measure of human difficulty. This dataset is central to our experiments as it uniquely allows us to compare the representations of human and model perceived difficulty on the same set of questions.

\textbf{Metrics.} To evaluate the performance of our probes, we report the Spearman rank correlation for the linear regression probe used on the expected success rate, $\hat{s}_{MC}$. The rank ordering of questions by difficulty is often a key requirement for downstream tasks such as difficulty-routing, see \Cref{sec:routing}. For binary success predictions, we report the AUROC, which measures discrimination quality irrespective of cutoff.

\textbf{Baselines.} Alongside the linear probes described in \Cref{sec:activation probes} we also evaluate a naive question-length baseline, a TF-IDF linear model, and a more complex MLP probe. We provide full details for all of these models in  Appendix \ref{app:Probing_Formulation}.

\subsection{Results}
\begin{table}[h!]
\centering
\caption{\textbf{Human and Model notions of difficulty are both linearly represented within the model activations.}
We report the Spearman correlation coefficient for probes trained on the E2H-AMC Dataset with both human IRT
and expected success rate labels. We note that: i) For both notions of difficulty the linear probe outperforms the TF-IDF
and length baselines; ii) The Human IRT difficulty is more linearly accessible within the model than expected success rate difficulty;
iii) the expected success rate probe performance monotonically decreases as the reasoning level of GPT-OSS increases.}

\label{tab:difficulty_encoding_comparison}
\small
\sisetup{
  table-number-alignment=center,
  table-format=1.2,
  detect-weight=true,
  detect-inline-weight=math
}
\setlength{\tabcolsep}{10pt}
\begin{tabular}{l S S S S S}
\toprule
& \multicolumn{2}{c}{\textbf{Qwen2.5-Math}} & \multicolumn{3}{c}{\textbf{GPT-OSS-20B}} \\
\cmidrule(lr){2-3} \cmidrule(lr){4-6}
\textbf{Method} & {\textbf{1.5B}} & {\textbf{7B}} & {\textbf{Low}} & {\textbf{Med}} & {\textbf{High}} \\
\midrule
\multicolumn{6}{l}{\textit{Human IRT difficulty} $b(q)$} \\
Linear Probe    & \textbf{0.85} & \textbf{0.87} &\textbf{0.84} & \textbf{0.83} & \textbf{0.83} \\
MLP Probe       & 0.83 & \textbf{0.87} & 0.77 & 0.79 & 0.80 \\
TF--IDF  & 0.72 & 0.72 & 0.74 & 0.74 & 0.74 \\
Length  & 0.15 & 0.15 & 0.15 & 0.15 & 0.15 \\
\midrule
\addlinespace[4pt]
\multicolumn{6}{l}{\textit{Model difficulty} $\hat{s}_{\mathrm{MC}}$} \\
Linear Probe    & 0.64 & \textbf{0.64} & \textbf{0.58} &	\textbf{0.50} &	\textbf{0.40} \\
MLP Probe    & \textbf{0.65} & 0.61 & 0.44 & 0.44 & 0.11 \\
TF--IDF  & 0.47 & 0.47 & 0.42 &	0.31 &	0.25 \\
Length  & 0.27 & 0.27 & 0.30 & 0.24 & 0.19 \\

% human
% qwen 1.5b => 0.834841453937175
% qwen 7b => 0.866152232590152
% low => 0.7671693520540174
% medium =>0.7940214797625322
% high => 0.801545487670544

% model success rate
% low => 0.4133547693950035
% medium => 0.43789170586721365
% high => 0.10465823955813654

\bottomrule
\end{tabular}
\end{table}

\paragraph{Human and model difficulty are both linearly decodable but encode different information.}
Table~\ref{tab:difficulty_encoding_comparison} shows that linear probes can predict both human IRT difficulty and model success rates from identical pre-generation activations, but with different levels of accessibility.
Human difficulty is consistently more linearly decodable (Spearman $\rho = 0.83$--$0.87$) than model success rate ($\rho = 0.40$--$0.64$) across all models, suggesting that models robustly encode what humans find difficult, even when that differs from their own performance characteristics.

Critically, the model success rate becomes \emph{less recoverable} as reasoning capability increases.
For GPT-OSS-20B, probe performance drops from $\rho = 0.58$ (low reasoning) to $\rho = 0.40$ (high reasoning), despite higher task success rate.
Non-linear MLP probes do not recover this loss, and in fact degrade more rapidly than linear probes under increased reasoning.
Together, these results suggest that while human-aligned difficulty is stably encoded, model-specific success signals are more fragile and sensitive to inference-time computation.
%This suggests that extended chain-of-thought may encode difficulty information in ways that are not linearly separable at the pre-generation stage, foreshadowing the routing challenges we address in Section~\ref{sec:routing}.

%Probe performance (AUROC) and task accuracy across models and inference regimes in Math and Coding domains. \textbf{Task Acc.} shows the model's average benchmark performance. \textbf{Math}: AUROC averaged over MATH, GSM8K, and AIME-2025, comparing Greedy vs.\ Maj@5. For GPT-OSS-20B we fix Maj@5 and vary the internal reasoning levels. \textbf{Code}: LiveCodeBench with target Pass@5 (a problem is correct if any of 5 sampled generations passes all test cases). Per-dataset results are reported in Appendix~\ref{tab:detailed_benchmark_performance}. \will{whilst info on metrics is here, it can be presented clearer. Why are we showing the Task Acc?}

\begin{table}[h!]
\caption{\textbf{We report the AUROC for probes and baselines trained on the binary notion of difficulty under different 
 decoding procedures}. The results are grouped by task: \textit{Math} or \textit{Code}, and decoding regime: \textit{Maj@5}, \textit{Greedy}, and \textit{Pass@5}.
\textit{Math} results show the AUROC averaged over the MATH, GSM8K, and AIME-2025 datasets, \textit{Code} reports the AUROC on 
LiveCodeBench with correctness defined w.r.t Pass@5 success. We also report the task success rate (Task Succ.) for each model 
to provide context on the number of positive and negative samples in training each probe.}
\label{tab:probe_inference_regimes}
\resizebox{\textwidth}{!}{
\begin{tabular}{llccccc}
\toprule
Model & Inference Regime & Task Succ. $\uparrow$ & Linear $\uparrow$ & MLP $\uparrow$ & TF-IDF $\uparrow$ & Length $\uparrow$ \\
\midrule
\multicolumn{7}{l}{\textit{Math: Greedy vs. Maj@5}} \\
\midrule
\multirow{2}{*}{Qwen2.5-Math-1.5B}
& Greedy & 0.724 & 0.84 & 0.83 & 0.64 & 0.61 \\
& Maj@5 & 0.763 & 0.76 & 0.75 & 0.63 & 0.66 \\
\cmidrule{1-7}
\multirow{2}{*}{Qwen2.5-Math-7B}
& Greedy & 0.809 & 0.79 & 0.79 & 0.68 & 0.67 \\
& Maj@5 & 0.827 & 0.80 & 0.77 & 0.72 & 0.66 \\
\cmidrule{1-7}
\multirow{2}{*}{Qwen2.5-1.5B}
& Greedy & 0.525 & 0.68 & 0.75 & 0.63 & 0.73 \\
& Maj@5 & 0.583 & 0.85 &  0.77 & 0.65 & 0.69 \\
\midrule
\multicolumn{7}{l}{\textit{Math: Maj@5 with Variable Reasoning Budget}} \\
\midrule
\multirow{3}{*}{GPT-OSS-20B}
& Reasoning: Low    & 0.866 & 0.78 & 0.79 & 0.68 & 0.62 \\
& Reasoning: Medium & 0.914 & 0.70 & 0.67 & 0.63 & 0.55 \\
& Reasoning: High   & 0.920 & 0.64 & 0.76 & 0.58 & 0.46 \\
\midrule
\multicolumn{7}{l}{\textit{Code: LiveCodeBench (target = Pass@5)}} \\
\midrule
Qwen2.5-Coder-3B & Pass@5 & 0.14 & 0.91 & 0.56 & 0.86 & 0.64 \\
Qwen2.5-Coder-7B & Pass@5 & 0.15 & 0.90 & 0.89 & 0.84 & 0.61 \\
DeepSeek-R1-Distill-Qwen-7B & Pass@5 & 0.30 & 0.81 & 0.80 & 0.83 & 0.59 \\
GPT-OSS-20B (low) & Pass@5 & 0.77 & 0.71 & 0.65 & 0.66 & 0.64 \\
GPT-OSS-20B (medium) & Pass@5 & 0.77 & 0.67 & 0.59 & 0.64 & 0.60 \\
GPT-OSS-20B (high) & Pass@5 & 0.79 & 0.69 & 0.64 & 0.64 & 0.62 \\
\bottomrule
\end{tabular}
}
\end{table}

\begin{figure}[h!]
    \centering
    \includegraphics[width=.8\linewidth]{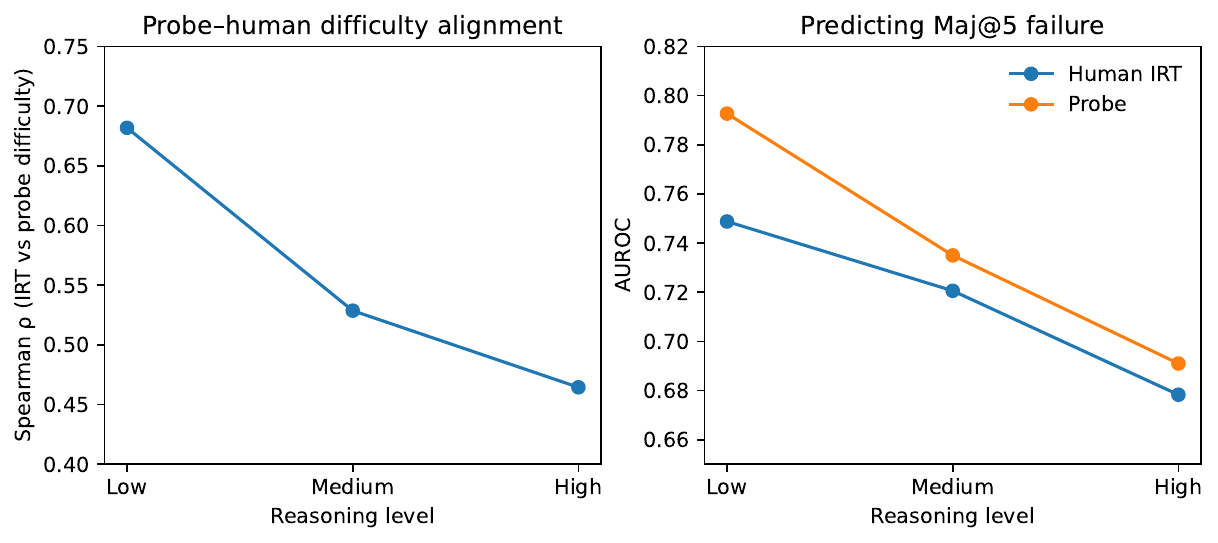}
    \caption{\textbf{Human and model difficulty diverge with increased reasoning.} We analyse how reasoning efforts affect the probe-able notions of human and binary difficulty on the E2H-AMC dataset using the GPT-OSS model. \textbf{Left:} We plot the Spearman correlation between IRT difficulty scores and the probe predicted difficulty across reasoning levels. As the reasoning level increases the correlation between probe-predicted model difficulty and human IRT difficulty labels decreases reflecting the diverging capabilities of the student examinees and model capabilities at higher reasoning levels. \textbf{Right:} Here we plot the AUROC of the probe and human IRT ground truth labels, which reflect probability of success on a question, against the model success at Maj@5, across reasoning levels. The difference between the human and probe scores indicates the probe is likely using an internal representation of model difficulty different from the human scores.}
    \label{fig:human_model_divergence}
\end{figure}

\paragraph{Binary success under specified decoding policies is more predictable than success rate.}
While success-rate prediction shows moderate correlation (Table~\ref{tab:difficulty_encoding_comparison}), binary classification of success under fixed decoding policies achieves substantially stronger discrimination.
Table~\ref{tab:probe_inference_regimes} shows that probes predicting Maj@5 or greedy success achieve AUROC $> 0.7$ across most settings, with several exceeding 0.8.

We observe three key patterns:
\begin{enumerate}[leftmargin=*, itemsep=2pt]
    \item \textbf{Greedy vs. sampling}: Greedy decoding generally yields higher probe AUROC than Maj@5 for the same model (e.g., Qwen2.5-Math-1.5B: 0.84 vs 0.76), likely because deterministic generation reduces noise in the prediction target.
    
    \item \textbf{Model capability matters}: Smaller or less capable models (e.g., Qwen2.5-1.5B, base variant) show stronger probe performance for Maj@5 than greedy, suggesting that sampling-based aggregation helps models solve problems they find marginally difficult, and this regime is easier to predict.
    
        \item \textbf{Reasoning budget degrades probe quality}: For GPT-OSS-20B, increasing the reasoning level from low to high decreases AUROC from 0.78 to 0.64 even under fixed Maj@5 decoding. However, unlike in the success-rate setting, this degradation is partially recoverable with non-linear probes: MLPs match or exceed linear probe performance at higher reasoning levels (e.g., 0.76 vs 0.64 at high reasoning). This suggests that while extended reasoning introduces representation drift, the underlying success signal remains present but becomes less linearly separable.
\end{enumerate}

% \paragraph{Code domain shows high probe quality.}
% On LiveCodeBench with Pass@5 as the target, we observe strong probe performance (AUROC $= 0.81$--$0.91$) for Qwen2.5-Coder and DeepSeek-R1 models.
% However, GPT-OSS-20B again shows weaker probe quality (AUROC $\approx 0.67$), consistent with the pattern in math domains.
% This suggests that probe accessibility is a model-family property that generalizes across domains rather than task-specific.% phenomenon.

\paragraph{Human and model difficulty diverge with increased reasoning.}
Figure~\ref{fig:human_model_divergence} illustrates how the relationship between human and model difficulty changes as GPT-OSS-20B's reasoning budget increases.
\Cref{fig:human_model_divergence} [Left] shows that alignment between probe-predicted model difficulty and human IRT difficulty decreases monotonically with reasoning level (Spearman $\rho$ drops from $\sim$0.65 to $\sim$0.45).
This indicates that as models become better at solving human-hard problems through extended reasoning, the notions of difficulty diverge. %s from human judgments.

\Cref{fig:human_model_divergence} [Right] demonstrates that despite this divergence, probe-based predictions of model difficulty consistently outperform human difficulty for predicting Maj@5 failures across all reasoning modes.
This confirms that models encode a model-relative notion of difficulty that is distinct from, and more predictive of their own performance than, human difficulty.

\begin{figure}[h!]
    \centering
    \includegraphics[width=\linewidth]{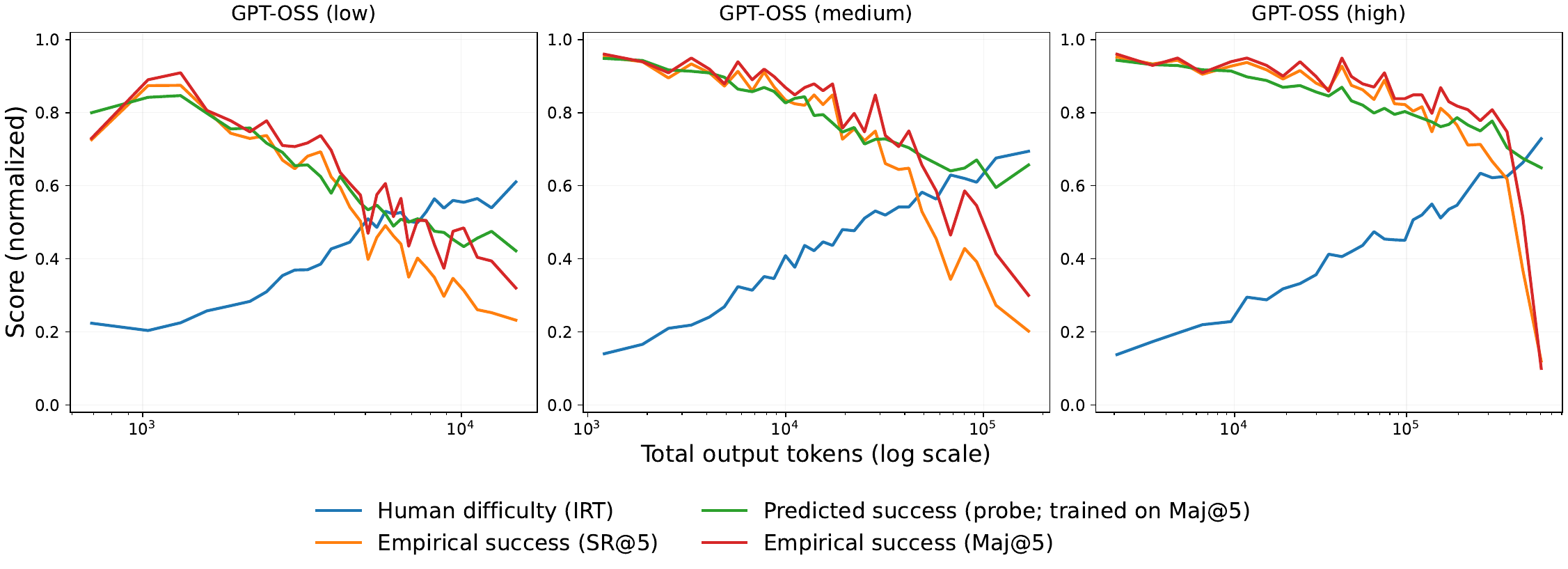}
\caption{\textbf{Chain-of-thought length tracks human difficulty but diverges from model success.}
On E2H-AMC, we plot binned chain-of-thought length (total output tokens, log-scale) against mean normalized human IRT difficulty, empirical success, and probe-predicted success (SR@5 and Maj@5) for GPT-OSS-20B across low, medium, and high reasoning modes. Across all settings, output length increases with human difficulty but decreases with both empirical and predicted success. This effect strengthens with higher reasoning budgets, indicating that generation length increasingly reflects human-aligned difficulty rather than model-relative likelihood of failure.}
    \label{fig:divergence_cot_length}
\end{figure}
\paragraph{Reasoning length reflects human difficulty rather than model uncertainty.
} To understand why human and model difficulty decouple under extended reasoning, we examine how chain-of-thought length (total output tokens) relates to human difficulty, empirical success, and probe-predicted model success across reasoning budgets. Figure~\ref{fig:divergence_cot_length} shows that as reasoning depth increases, output length becomes increasingly correlated with human IRT difficulty, while simultaneously becoming negatively-correlated with both empirical success and probe-predicted success. For GPT-OSS, this pattern is consistent across reasoning modes and strengthens at higher budgets: the model spends more tokens on problems humans find difficult, even when those problems are well within the model’s competence. This observation aligns with concurrent work by \citet{chen_think_2026}, which finds that longer chain-of-thought traces are not a reliable indicator of correctness. Our results provide a complementary perspective: we show that this disconnect arises because reasoning length tracks human-aligned difficulty, while model-relative success remains predictable from pre-generation representations.

As such, extended reasoning amplifies a human-aligned difficulty signal %in generation dynamics
that is distinct from the model’s own likelihood of success, helping explain why probe-predicted model difficulty remains useful even as alignment with human difficulty deteriorates.

\section{Practical Applications: Probe-Guided Routing}
\label{sec:routing}
Prior work on routing between models with different capabilities and inference costs 
typically relies on indirect proxies for difficulty, such as input length, perplexity, 
or heuristic confidence measures~\cite{chen_frugalgpt_2024, ding_hybrid_2023}. 
We demonstrate that \emph{probe-derived success estimates enable effective routing decisions}, 
yielding meaningful performance-cost tradeoffs in both cascade and utility-based settings.

\label{sec:routing_results}
\begin{figure}[h!]
    \centering
    \includegraphics[width=\linewidth]{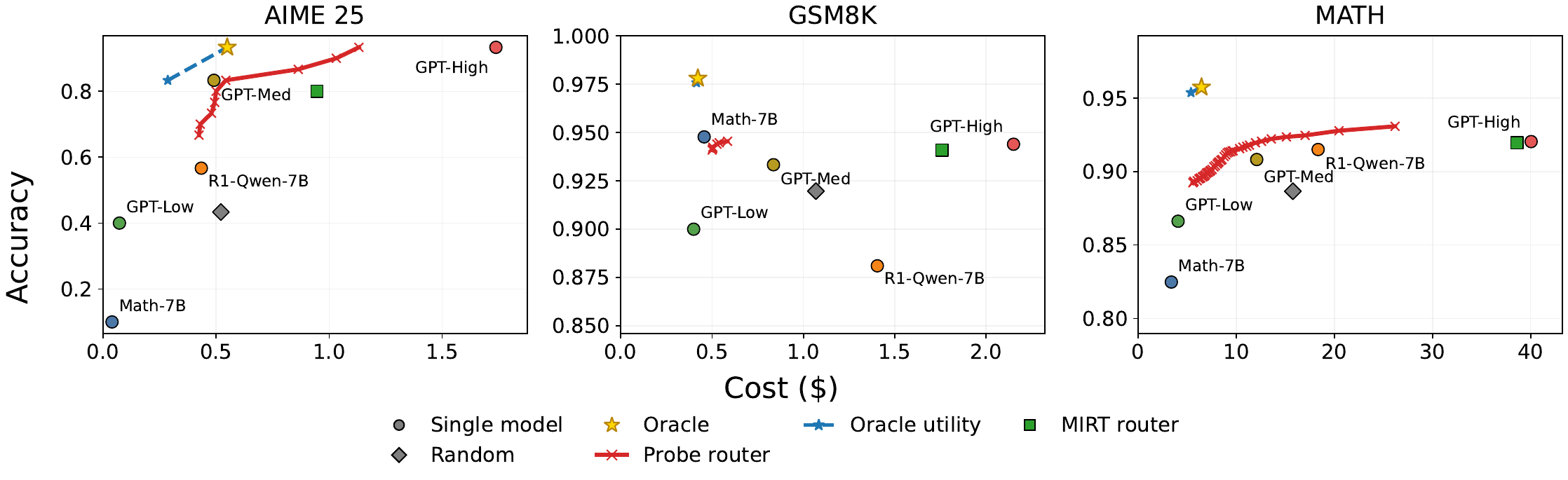}
\caption{\textbf{Probe-based routing generalizes across diverse reasoning benchmarks.}
On the hardest benchmark (AIME 2025), the router matches GPT-OSS-20B-high's 93.3\% 
accuracy at 37\% lower cost, tracing a Pareto frontier that dominates all single-model and IRT Router
baselines. On the saturated benchmark (GSM8K), it pivots to cost minimization, 
identifying Math-7B as the cost-optimal choice and avoiding expensive models that offer 
no accuracy gain. On the intermediate benchmark (MATH), it achieves a Pareto improvement 
over all baselines, matching GPT-OSS-20B-high accuracy while reducing cost by 
approximately 70\%. Together, these results show that a single probe-based routing 
mechanism adapts to difficulty distributions across benchmarks without retraining. All 
results use maj@5 with $K{=}5$ generations. For full results, see 
Appendix~\ref{app:routing_strategy_results}.}
\label{fig:routing_generalization}
\end{figure}

% \begin{figure}[h!]
%     \centering
%     \begin{minipage}{0.48\linewidth}
%         \centering
%         \includegraphics[width=\linewidth]{figs/pareto_cascade_MATH_7b_gpt_20b.png}
%     \end{minipage}
%     \hfill
%     \begin{minipage}{0.48\linewidth}
%         \centering
%         \includegraphics[width=\linewidth]{figs/DigitalLearningGmbH_MATH-lighteval_pareto_with_mirt_single.png}
%     \end{minipage}
%     \caption{\textbf{Probe-based routing achieves strong performance-cost tradeoffs on MATH.} 
% \textbf{Left (Cascade):} Binary routing between Qwen2.5-Math-7B-Instruct and GPT-OSS-20B-medium. 
% The cascade strategy (red curve) substantially outperforms random routing (gray diamond) across the Pareto frontier, matching GPT-OSS-20B-medium accuracy (orange circle) at 17\% lower cost. 
% \textbf{Right (Utility):} Model selection from a pool of five models with varying capabilities and costs. 
% The utility router (red curve) achieves a Pareto improvement over all single-model baselines, exceeding GPT-OSS-20B-high accuracy (red circle) while reducing cost by approximately 70\%. 
% %Both strategies route difficult queries (low $\hat{p}$) to more capable models. 
% Oracle performance (gold star) represents an upper bound with perfect difficulty prediction. 
% All results use maj@5 with $K=5$ generations.}
% \label{fig:routing_main}
% \end{figure}

\subsection{Routing Strategies}
\label{sec:routing_setup}

We evaluate a probe-based utility routing strategy that uses predicted success probabilities 
$\hat{p}_M(x)$ to allocate queries across a set of models $M$, with different capabilities and costs. A cascading strategy is explored within the Appendix \ref{app:Cascade_Approach}.

\paragraph{Utility-Based Probe Routing}
\label{sec:utility_routing}

For routing among a heterogeneous pool of models, we use a simple utility-based rule. Let $\{M_1, \ldots, M_K\}$ denote available models with expected costs $\{\hat{c_1}, \ldots, \hat{c_K}\}$ based on the average output cost from the training set. We normalize the expected cost $\hat{c_i}$ to be between [0-1] such that for a prompt $x$, we select:
\[
\hat{M}(x) = \arg\max_i \left( \hat{p}_i(x) - \lambda \hat{c_i} \right)
\]
where $\hat{p}_i(x)$ is the probe-estimated success probability for model $M_i$, 
and $\lambda$ trades off success probability against cost. This requires training 
separate probes for each model in the pool.

\subsection{Routing Experiment Setup}

% We evaluate on a heterogeneous pool of models spanning sizes and reasoning capabilities: GPT-OSS-20B (low/medium/high reasoning)~\citep{openai2025gptoss}, DeepSeek-R1-Distill-Qwen-7B~\citep{deepseekai2025deepseekr1}, and Qwen2.5-Math (1.5B, 7B) \citep{yang2024qwen25math}. Full generation configurations are provided in Appendix~\ref{sec:model_settings}.

We evaluate using a pool of five models: Qwen2.5-Math-7B-Instruct \citep{yang2024qwen25math}, Deepseek-R1-Qwen-7B \citep{guo2025deepseek}, and GPT-OSS-20B \citep{openai2025gptoss} with low/medium/high reasoning budgets. We vary $\lambda$ to trace the performance-cost frontier. %, without tuning for specific operating points. 
Following prior routing work that uses API pricing to estimate deployment costs \citep{chen_frugalgpt_2024, ding_hybrid_2023}, we use Fireworks AI's inference pricing to emulate realistic conditions, %-- a platform that provides realistic cost estimates for deploying open-source models at scale 
see Appendix~\ref{app:routing_strategy_results}. All experiments use maj@5 accuracy with K=5 generations on the MATH~\cite{hendrycksmath2021} dataset, and probes trained to predict maj@5 success as described in Section~\ref{sec:predicting_difficulty}.

\paragraph{Baselines}
We compare against three baselines: an IRT router baseline, random routing, and an oracle with perfect knowledge of model success. Random routing assigns each problem uniformly at random to one of the available models, independent of difficulty or cost. The IRT router follows prior work on modeling question difficulty and model ability via latent trait estimation using learned embeddings \citep{song_irt-router_2025}. Unlike our approach, which directly extracts model-specific success signals from internal representations, IRT-based routers rely on learned item and model embeddings and require additional training to estimate these latent variables.

For utility-based routing, the oracle replaces probe predictions with ground-truth correctness labels $(p(x)=[\text{correct}_i(x)])$ and  selects $\hat{M}(x) = \arg\max_i (\mathbb{I}[\text{correct}_i(x)] - \lambda \hat{c_i})$ sweeping $\lambda$ to trace the theoretical best-case Pareto frontier.

We report additional routing configurations, including the Self-Consistency(SC)-Entropy baseline and routing considering input-prompt costs, in
Appendix~\ref{app:full_cost_routing}.

\subsection{Results}

\paragraph{Utility routing achieves strong cost-accuracy tradeoffs and adapts to benchmark difficulty.}
Figure~\ref{fig:routing_generalization} shows that utility routing yields strong gains across benchmarks, achieving substantial cost reductions while maintaining accuracy. On MATH, the router matches GPT-OSS-20B-high's 92\% accuracy at a 70\% cost reduction. This behavior generalizes across benchmarks with different difficulty distributions. On AIME 2025, where model performance varies widely (40\%–93\%), the router matches the strongest model's performance at a 37\% cost reduction (\$1.15 vs \$1.75). In contrast, on GSM8K, where performance saturates across models (85\%–95\%), the router identifies the cost-optimal model, selecting Math-7B (94.5\% at \$0.34) over significantly more expensive high-reasoning models (GPT-OSS-20B-high: 94.4\% at \$2.4). These results demonstrate performance-aware allocation: routing preferentially uses stronger models when task difficulty varies, and shifts toward efficient models when accuracy plateaus. See Appendix~\ref{app:routing_strategy_results} for full results.

\subsection{Discussion}
\label{sec:discussion}
\paragraph{Reasoning changes what difficulty means, not just performance}
Our results show that increased test-time reasoning fundamentally changes how difficulty is represented in LLMs. While extended reasoning improves task success rate, it consistently reduces the linear accessibility of pre-generation success signals. Across reasoning modes in GPT-OSS-20B, the linear probe AUROC drops monotonically as reasoning budgets increase, even as accuracy improves. Analysis of chain-of-thought length reveals a key mechanism: with deeper reasoning, generation length becomes increasingly correlated with human difficulty rather than the model’s own likelihood of failure. As a result, reasoning traces amplify human-aligned difficulty signals that decouple from model-relative uncertainty, explaining why probes degrade precisely when reasoning is most effective.

\paragraph{Human difficulty and model difficulty are distinct—and diverge with capability}The divergence between human and model difficulty has broader implications beyond routing. Using E2H-AMC, we show that LLMs robustly encode human psychometric difficulty even when that signal no longer predicts model failure. As reasoning capability increases, models increasingly solve problems that humans find difficult, yet their internal representations continue to track human-aligned difficulty through longer reasoning traces. This creates a growing mismatch: human difficulty remains linearly accessible, while model-relative difficulty becomes harder to extract during extended reasoning. For applications such as curriculum learning, data selection, or evaluation, this suggests that human difficulty labels may increasingly mischaracterise what models actually find challenging.

\paragraph{Routing effectiveness is mediated by probe reliability.}

Probe-guided routing approaches oracle-utility performance when probes achieve high discrimination (AUROC), but exhibits substantial gaps when probe quality degrades. This suggests routing effectiveness is constrained by the reliability of success estimates rather than model capability alone. Even in lower-quality regimes, the probe consistently identifies cost-effective models, selecting cheaper models on saturated benchmarks (GSM8K) and higher-capability models on harder tasks (AIME). Unlike embedding-based approaches such as IRT router \citep{song_irt-router_2025}, which depend on learned latent representations from another model, probe-based routing from the very model whose performance it is predicting.

\section{Conclusion and Limitations}

We show that a model's likelihood of success is encoded before generation begins and can be extracted with simple linear probes. These signals reflect a notion of difficulty that differs from human judgments but more reliably predicts model performance. While they generalize across decoding strategies, they become harder to access as test-time compute increases. When accessible, they enable routing strategies that approach oracle performance.

\textbf{Limitations.} We focus on linear probes at a single post-instruction position. 
While effective for base and lightly instruction-tuned models, probe performance degrades 
under extended reasoning, and we do not probe during generation or do cross-domain transfer (e.g., math to code). Our probes are sensitive to token position, and our routing policies use fixed-$k$ majority voting rather than learned or adaptive selection of $k$.

\section{Acknowledgements}
WL is supported by the Rhodes Scholarship. WB was supported by the Engineering and Physical Sciences Research Council EP/S021566/1. Lastly, we thank the reviewers for their helpful feedback.

\bibliography{colm2026_conference}
\bibliographystyle{colm2026_conference}

\appendix
\section{Probe and Baseline Formulation}
\label{app:Probing_Formulation}

\paragraph{Linear probe.}
Let $h_i^{\ell} \in \mathbb{R}^D$ denote the residual-stream hidden state at layer $\ell$ and token position $i$ from the frozen language model. These are the internal representations we probe.

\paragraph{Finding end-of-instruction positions.} We identify where instructions end by applying the model's chat template to a placeholder input, then tokenizing the post-instruction suffix. This gives us $P$ token positions relative to the last non-padding token: $\{-P, \ldots, -1\}$. These are our candidate positions to probe, see \Cref{fig:probe suffix positions}.
% \vspace{-0.1cm}
% \begin{tcolorbox}[colback=gray!10, colframe=gray!50, boxrule=0.5pt, boxsep=2pt, top=3pt, bottom=3pt]
% \small
% \textbf{Example suffixes:}\\[3pt]
% \textbf{Qwen2.5:} \texttt{<|im\_end|>\textbackslash n<|im\_start|>assistant\textbackslash n} \\
% \textbf{DeepSeek-R1:} \texttt{<|Assistant|><think|>\textbackslash n} \\
% \textbf{GPT-OSS:} \texttt{<|end|><|start|>assistant}
% \end{tcolorbox}
% \vspace{-0.1cm}

\paragraph{Training linear probes.} For each candidate pair $(\ell, p) \in \mathcal{L} \times \mathcal{P}$, where $\mathcal{L}$ spans all transformer layers and $\mathcal{P}$ the post-instruction suffix positions, we train a single linear probe on the activation vector $h_p^{\ell}$. The probe has no bias term, just a linear map from the $D$-dimensional activation to predictions.

The task type is determined automatically: for continuous success-rate targets we use Ridge regression (evaluated with Spearman's $\rho$); for binary correctness labels we use $\ell_2$-regularized logistic regression (evaluated with ROC-AUC). The regularisation strength $\alpha$ is tuned via grid search on validation data over the range $\alpha \in \{10^{-3}, 10^{-2}, 10^{-1}, 1, 10, 10^2, 10^3, 10^4\}$.

\paragraph{Data and evaluation.} We hold out 20\% of the original training set as validation. The best configuration $(\ell^*, p^*, \alpha^*)$ is selected by validation performance. For classification, we apply Platt scaling on validation data to calibrate probabilities. Test evaluation happens exactly once with the selected probe.

\paragraph{MLP probes.} While our main results use linear probes for efficiency and interpretability, we also implement MLP-based probes to capture non-linear relationships in internal representations.

For each candidate pair $(\ell, p) \in \mathcal{L} \times \mathcal{P}$, we train a small 2-layer MLP on the activation vector $h_p^{\ell} \in \mathbb{R}^D$. The architecture consists of:
\begin{itemize}
    \item \textbf{Input layer:} $D$ dimensions (matching model hidden dimension)
    \item \textbf{Hidden layer:} 256 dimensions with ReLU activation
    \item \textbf{Output layer:} 1 dimension (for regression or binary classification)
\end{itemize}

For regression tasks (success-rate prediction), we use MSE loss with output activation $\text{identity}$. For binary classification (majority vote correctness), we use binary cross-entropy loss with sigmoid output activation.

\paragraph{MLP training details.} We apply $\ell_2$ regularization with strength $\alpha$ tuned via grid search: $\alpha \in \{10^{-4}, 10^{-3}, 10^{-2}, 10^{-1}, 1\}$. Optimisation is performed with Adam (learning rate $10^{-3}$) for up to 100 epochs with early stopping on validation loss (patience=10). Data is split 80/20 for train-validation. For classification, we apply Platt scaling on validation data to calibrate probabilities. Best configuration $(\ell^*, p^*, \alpha^*)$ is selected by validation performance and evaluated exactly once on a held-out test set.

\paragraph{TF--IDF probe.}
Using spaCy, we preprocess each prompt with lemmatisation and stop-word removal, then represent it using TF--IDF features. We train Ridge regression for continuous targets and logistic regression for binary targets. We select the regularisation strength on the validation set using the same grid and metrics as the linear probes, then evaluate once on the test set.

\paragraph{Length probe.}
We represent each prompt using a single feature: its whitespace-separated word count. We train Ridge regression for continuous targets and logistic regression for binary targets. Regularisation is selected on the validation set using the same procedure as the TF--IDF probe, followed by evaluation on a held-out test set.

\paragraph{Comparison.} Comparing linear and MLP
probes, three patterns emerge: (1)~for human IRT difficulty 
prediction, the two probe types perform comparably, suggesting 
this signal is robustly linear; (2)~for binary success prediction 
under standard decoding, MLP probes offer marginal or no improvement; 
(3)~under variable reasoning with GPT-OSS, MLP probes partially recover 
discrimination lost by linear probes (e.g., AUROC $0.76$ vs $0.64$ 
for GPT-OSS-20B-high on Maj@5), suggesting that reasoning transforms 
representations in ways that break linear separability while 
preserving the signal in a nonlinear subspace. However, this 
recovery is task-dependent: for continuous success-rate regression, 
MLP probes degrade more severely than linear probes at high 
reasoning (Table~\ref{tab:difficulty_encoding_comparison}), indicating that the 
underlying signal may become genuinely less accessible rather 
than merely nonlinear.

\newpage

\section{Dataset Statistics}
\label{app:dataset statistics}

\begin{table}[h]
\centering
\caption{Train, validation, and test split sizes for each dataset.}
\label{tab:dataset-splits}
\begin{tabular}{lrrr}
\toprule
Dataset & Train & Val & Test \\
\midrule
MATH           & 6{,}000 & 1{,}500 & 5{,}000 \\
E2H-AMC        &     640 &     160 & 2{,}975 \\
AIME-1983--2024 &    598 &     149 &      30 \\
AIME-2025      &      -- &      -- &      30 \\
GSM8K          & 4{,}784 & 1{,}195 & 1{,}319 \\
LiveCodeBench  &     458 &     114 &     341 \\
\bottomrule
\end{tabular}
\end{table}

\section{Probe Layer Sweep}
\label{app:probe layer sweep}

\begin{figure}[h]
    \centering
    \begin{subfigure}[t]{0.48\textwidth}
        \centering
        \includegraphics[width=\linewidth]{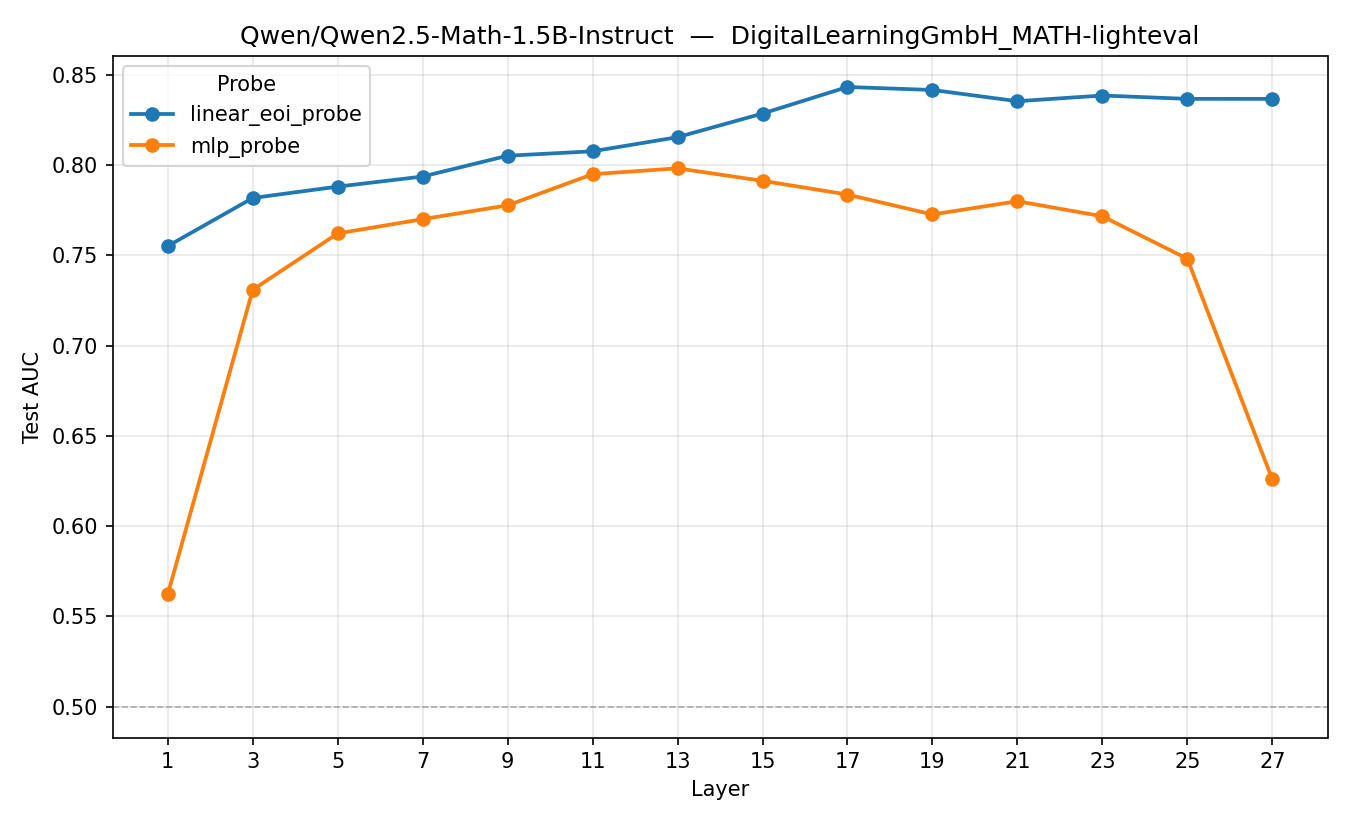}
        %\caption{Validation accuracy across probe layers.}
        \label{fig:left}
    \end{subfigure}
    \hfill
    \begin{subfigure}[t]{0.48\textwidth}
        \centering
        \includegraphics[width=\linewidth]{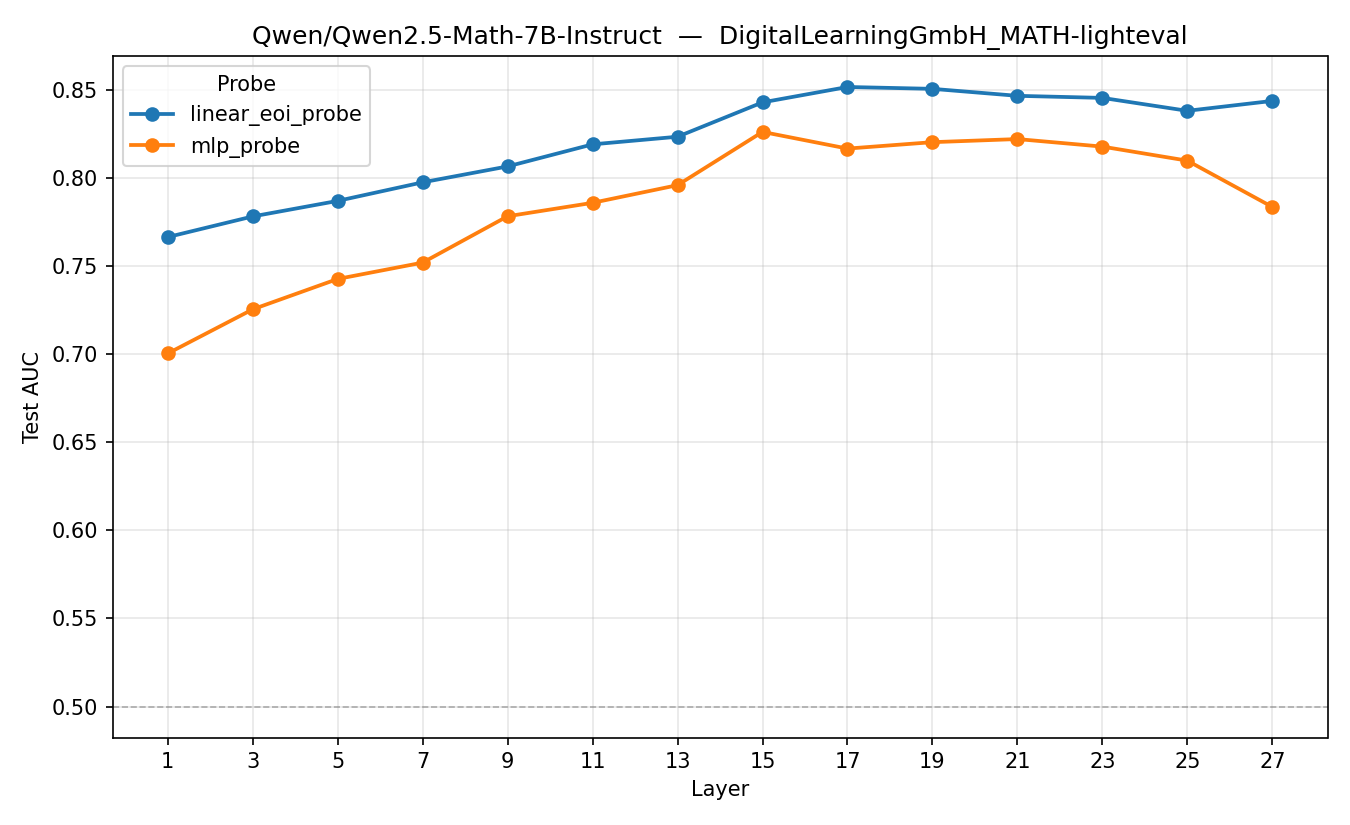}
        %\caption{}
        \label{fig:right}
    \end{subfigure}
    \caption{The validation performance of the linear and MLP probes at different layers for the \textsc{Qwen2.5-Math-1.5B-Instruct} \subref{fig:left} and \textsc{Qwen2.5-Math-7B-Instruct} \subref{fig:right} models on the MATH dataset. In both models the linear probe outperforms the MLP probe. The linear probe performance tends to increase as the layer index increases.}
    \label{fig:main}
\end{figure}

\begin{table}[t]
\centering
\caption{Best probe layer and token position selected on validation, with the corresponding validation and test scores. Note the best token position is indexed backward from the final token.}
\label{tab:probe-sweep}
\begin{tabular}{llccrr}
\toprule
Model & Dataset & Layer & Pos. & Val score & Test score \\
\midrule
\multirow{4}{*}{Qwen2.5-Math-7B-Instruct}
  & MATH  & 21 & 4 & 0.891 & 0.846 \\
  & AMC   & 22 & 4 & 0.881 & 0.848 \\
  & AIME  & 28 & 1 & 0.867 & 0.679 \\
  & GSM8K & 20 & 4 & 0.754 & 0.767 \\
\midrule
\multirow{3}{*}{DeepSeek-R1-Distill-Qwen-7B}
  & MATH  & 27 & 2 & 0.793 & 0.739 \\
  & AIME  & 13 & 1 & 0.753 & 0.701 \\
  & GSM8K & 23 & 1 & 0.647 & 0.640 \\
\midrule
\multirow{4}{*}{gpt-oss-20b (high)}
  & MATH  & 15 & 0 & 0.850 & 0.860 \\
  & AMC   & 16 & 2 & 0.757 & 0.691 \\
  & AIME  & 10 & 2 & 0.761 & 0.286 \\
  & GSM8K & 17 & 0 & 0.701 & 0.686 \\
\midrule
\multirow{4}{*}{gpt-oss-20b (low)}
  & MATH  & 17 & 0 & 0.845 & 0.845 \\
  & AMC   & 12 & 0 & 0.810 & 0.793 \\
  & AIME  & 24 & 0 & 0.825 & 0.778 \\
  & GSM8K & 17 & 0 & 0.790 & 0.761 \\
\midrule
\multirow{4}{*}{gpt-oss-20b (medium)}
  & MATH  & 15 & 0 & 0.840 & 0.850 \\
  & AMC   & 13 & 1 & 0.784 & 0.735 \\
  & AIME  & 19 & 2 & 0.778 & 0.568 \\
  & GSM8K & 16 & 1 & 0.921 & 0.629 \\
\bottomrule
\end{tabular}
\end{table}

\section{Additional Probe Results}
\label{app:Additional Probe Results}

\begin{table}[H]
\caption{AUROC for Predicting Accuracy under Different Decoding Strategies}
\label{tab:majk_results}
\centering
\small
\begin{tabular}{llcc}
\toprule
\multicolumn{4}{c}{\textbf{Linear Probe}} \\
\midrule
Decoding & Dataset & Qwen2.5-Math-1.5B & Qwen2.5-Math-7B \\
\midrule
\multirow{3}{*}{Maj@5} 
& MATH & 0.838 & 0.845 \\
& AIME & 0.712 & 0.765 \\
& GSM8K & 0.717 & 0.784 \\
\midrule
\multirow{3}{*}{Greedy} 
& MATH & 0.837 & 0.827 \\
& AIME & 0.911 & 0.753 \\
& GSM8K & 0.762 & 0.778 \\
\midrule
\multicolumn{4}{c}{\textbf{MLP Probe}} \\
\midrule
Decoding & Dataset & Qwen2.5-Math-1.5B & Qwen2.5-Math-7B \\
\midrule
\multirow{3}{*}{Maj@5} 
& MATH & 0.834 & 0.846 \\
& AIME & 0.712 & 0.703 \\
& GSM8K & 0.692 & 0.77 \\
\midrule
\multirow{3}{*}{Greedy}  
& MATH & 0.825 & 0.827 \\
& AIME & 0.911 & 0.753 \\
& GSM8K & 0.762 & 0.778 \\
\midrule
\multicolumn{4}{c}{\textbf{Tfidf Probe}} \\
\midrule
\multirow{3}{*}{Maj@5} 
& MATH & 0.771 & 0.761 \\
& AIME & 0.529 & 0.840 \\
& GSM8K & 0.582 & 0.568 \\
\midrule
\multirow{3}{*}{Greedy} 
& MATH & 0.762 & 0.760 \\
& AIME & 0.589 & 0.741 \\
& GSM8K & 0.581 & 0.537 \\
\midrule
\multicolumn{4}{c}{\textbf{Length Probe}} \\
\midrule
\multirow{3}{*}{Maj@5} 
& MATH & 0.70& 0.68 \\
& AIME & 0.66 & 0.72 \\
& GSM8K & 0.62 & 0.59 \\
\midrule
\multirow{3}{*}{Greedy} 
& MATH & 0.70 & 0.70 \\
& AIME & 0.83 & 0.66 \\
& GSM8K & 0.66 & 0.62 \\
\bottomrule
\end{tabular}
\end{table}

% QWEN MATH 1.5
% MATH => 0.8335114534514214, GSM8K => 0.6922380769967439, AIME => 0.7115384615384616 ## maj@5 => 0.7457626639955423
% MATH => 0.8252072824085196, GSM8K => 0.7622379084530013, AIME => 0.9107142857142858 ## pass@1 =>0.83

% QWEN MATH 7B
% % MATH => 0.8463186315542387, GSM8K => 0.77, AIME => 0.7037037037037036 ##maj@5 => 0.7733407784193141
% % MATH => 0.827451511407985, GSM8K => 0.7781027020704177, AIME => 0.7530864197530864 ##pass@1 => 0.7862135444104963

% QWEN 1.5
% MATH => 0.8441365702502802, GSM8K => 0.7617115987460815, AIME => 0.9310344827586207 ##maj@5 => 0.8456275505849941
% MATH => 0.8285521147293552, GSM8K => 0.7216084323503786, AIME => XX(0.71) ##pass@1 =>0.75

% low pass@5 (0.7876600486)
% MATH => 0.8317481403404869, GSM8K =>0.7627134870184574, AIME =>0.7685185185185185 

% medium pass@5 (0.6735068437)
% MATH => 0.8251203390651816, GSM8K =>0.6914001920094528 , AIME => 0.504

% high pass@5 (0.7629739978)
% MATH =>0.8217226943059495 , GSM8K =>0.6100564419841529 , AIME => 0.8571428571428571

\begin{table}[H]
\caption{Maj@5 Probe Performance Comparison across GPT-OSS-20B thinking modes}
\label{tab:probe_results}
\centering
\small
\begin{tabular}{llccc}
\toprule
\multicolumn{5}{c}{\textbf{Linear Probe}} \\
\midrule
Difficulty & Dataset & Low & Medium & High \\
\midrule
\multirow{4}{*}{gpt-oss-20b} 
& MATH-lighteval & 0.848 & 0.842 & 0.855 \\
& AMC & 0.793 & 0.735 & 0.691 \\
& AIME & 0.731 & 0.600 & 0.375 \\
& GSM8K & 0.767 & 0.660 & 0.680 \\
\midrule
\multicolumn{5}{c}{\textbf{MLP Probe}} \\
\midrule
Difficulty & Dataset & Low & Medium & High \\
\midrule
\multirow{4}{*}{gpt-oss-20b} 
& MATH-lighteval & 0.831 & 0.25 & 0.823 \\
& AMC & 0.793 & 0.735 & 0.691 \\
& AIME & 0.769 & 0.504 & 0.858 \\
& GSM8K & 0.763 & 0.691 & 0.610 \\
\midrule
\multicolumn{5}{c}{\textbf{Tfidf Probe}} \\
\midrule
\multirow{4}{*}{gpt-oss-20b} 
& MATH-lighteval & 0.771 & 0.781 & 0.783 \\
& AMC & 0.692 & 0.649 & 0.625 \\
& AIME & 0.602 & 0.528 & 0.339 \\
& GSM8K & 0.670 & 0.580 & 0.627 \\
\bottomrule
\end{tabular}
\end{table}

\begin{table}[H]
\caption{Benchmark Performance across Different Models}
\label{tab:detailed_benchmark_performance}
\centering
\small
\begin{tabular}{lllcc}
\toprule
\multicolumn{5}{c}{\textbf{GPT-OSS-20B Performance}} \\
\midrule
Difficulty & Dataset & Low & Medium & High \\
\midrule
\multirow{5}{*}{gpt-oss-20b} 
& MATH-lighteval & 0.866 & 0.914 & 0.920 \\
& AMC & 0.603 & 0.778 & 0.837 \\
& AIME & 0.620 & 0.913 & 0.963 \\
& GSM8K & 0.900 & 0.933 & 0.944 \\
& AIME2025 & 0.400 & 0.833 & 0.933 \\
\midrule
\multicolumn{5}{c}{\textbf{Other Models Performance}} \\
\midrule
Decoding & Model & Dataset & Maj@5 & Greedy \\
\midrule
\multirow{12}{*}{Comparison} 
& \multirow{4}{*}{\begin{tabular}[c]{@{}l@{}}Qwen2.5-1.5B\\-Instruct\end{tabular}} 
& MATH-lighteval & 0.583 & 0.525 \\
& & AIME & 0.072 & 0.059 \\
& & GSM8K & 0.758 & 0.687 \\
& & AIME2025 & 0.033 & 0.000 \\
\cmidrule(lr){2-5}
& \multirow{4}{*}{\begin{tabular}[c]{@{}l@{}}Qwen2.5-Math\\-1.5B-Instruct\end{tabular}} 
& MATH-lighteval & 0.763 & 0.724 \\
& & AIME & 0.320 & 0.278 \\
& & GSM8K & 0.855 & 0.835 \\
& & AIME2025 & 0.167 & 0.067 \\
\cmidrule(lr){2-5}
& \multirow{4}{*}{\begin{tabular}[c]{@{}l@{}}Qwen2.5-Math\\-7B-Instruct\end{tabular}} 
& MATH-lighteval & 0.827 & 0.809 \\
& & AIME & 0.340 & 0.281 \\
& & GSM8K & 0.945 & 0.937 \\
& & AIME2025 & 0.100 & 0.100 \\
\bottomrule
\end{tabular}
\end{table}

\begin{table}[h]
\centering
\caption{Cross-dataset transfer. Each probe is trained on one dataset and
evaluated on all three. Bold on the diagonal marks in-distribution evaluation.}
\label{tab:cross-dataset}
\small
\begin{tabular}{llrrrr}
\toprule
Model & Trained on & MATH & GSM8K & AIME & Avg.\ $\pm$ SD \\
\midrule
\multirow{3}{*}{Qwen2.5-1.5B-Instruct}
  & MATH  & \textbf{0.844} & 0.727 & 0.931 & $0.834 \pm 0.084$ \\
  & GSM8K & 0.796 & \textbf{0.762} & 1.000 & $0.853 \pm 0.105$ \\
  & AIME  & 0.783 & 0.709 & \textbf{0.931} & $0.808 \pm 0.092$ \\
\midrule
\multirow{3}{*}{Qwen2.5-Math-1.5B-Instruct}
  & MATH  & \textbf{0.841} & 0.730 & 0.654 & $\mathbf{0.742 \pm 0.077}$ \\
  & GSM8K & 0.760 & \textbf{0.758} & 0.538 & $0.685 \pm 0.104$ \\
  & AIME  & 0.740 & 0.605 & \textbf{0.712} & $0.686 \pm 0.058$ \\
\midrule
\multirow{3}{*}{Qwen2.5-Math-7B-Instruct}
  & MATH  & \textbf{0.846} & 0.720 & 0.901 & $\mathbf{0.822 \pm 0.076}$ \\
  & GSM8K & 0.794 & \textbf{0.767} & 0.877 & $0.813 \pm 0.047$ \\
  & AIME  & 0.745 & 0.640 & \textbf{0.679} & $0.688 \pm 0.043$ \\
\midrule
\multirow{3}{*}{DeepSeek-R1-Distill-Qwen-7B}
  & MATH  & \textbf{0.739} & 0.515 & 0.643 & $\mathbf{0.632 \pm 0.092}$ \\
  & GSM8K & 0.589 & \textbf{0.640} & 0.502 & $0.577 \pm 0.057$ \\
  & AIME  & 0.634 & 0.511 & \textbf{0.701} & $0.615 \pm 0.079$ \\
\midrule
\multirow{3}{*}{gpt-oss-20b (low)}
  & MATH  & \textbf{0.845} & 0.710 & 0.611 & $\mathbf{0.722 \pm 0.096}$ \\
  & GSM8K & 0.662 & \textbf{0.761} & 0.361 & $0.595 \pm 0.170$ \\
  & AIME  & 0.650 & 0.614 & \textbf{0.778} & $0.681 \pm 0.070$ \\
\midrule
\multirow{3}{*}{gpt-oss-20b (medium)}
  & MATH  & \textbf{0.850} & 0.725 & 0.504 & $\mathbf{0.693 \pm 0.143}$ \\
  & GSM8K & 0.540 & \textbf{0.629} & 0.504 & $0.558 \pm 0.053$ \\
  & AIME  & 0.558 & 0.577 & \textbf{0.568} & $0.568 \pm 0.008$ \\
\midrule
\multirow{3}{*}{gpt-oss-20b (high)}
  & MATH  & \textbf{0.860} & 0.692 & 0.679 & $\mathbf{0.744 \pm 0.082}$ \\
  & GSM8K & 0.656 & \textbf{0.686} & 0.214 & $0.519 \pm 0.216$ \\
  & AIME  & 0.594 & 0.522 & \textbf{0.286} & $0.467 \pm 0.132$ \\
\bottomrule
\end{tabular}
\end{table}

\begin{table}[h]
\centering
\caption{Probe performance across Qwen2.5-7B variants with different degrees of post-training.}
\label{tab:qwen-variants}
\begin{tabular}{lccc}
\toprule
Qwen2.5-7B & MATH & AIME & GSM8K \\
\midrule
Base                   & 0.7694 & --     & 0.6661 \\
Math                   & 0.8463 & 0.6790 & 0.7665 \\
Instruct               & 0.8375 & 0.7284 & 0.7766 \\
Reasoning (R1-Distill) & 0.7391 & 0.7014 & 0.6401 \\
\bottomrule
\end{tabular}
\end{table}

\newpage

\section{Optimal Model Settings}
\label{sec:model_settings}
All rollouts were performed using VLLM \cite{kwon2023efficient} with the following configurations:

\begin{table}[h!]
\centering
\small
\begin{tabular}{lccc}
\toprule
\textbf{Model} & \textbf{Max Length} & \textbf{Temperature} & \textbf{k} \\
\midrule
GPT-OSS-20B & 131,072 & 1.0 & 5 \\
DeepSeek-R1-Distill-Qwen-7B & 32,768 & 0.6 & 5 \\
Qwen2.5-Math-1.5B-Instruct & 3,000 & 0.7 & 5 \\
Qwen2.5-Math-7B-Instruct & 3,000 & 0.7 & 5 \\
Qwen2.5-Coder-XB-Instruct & 4,096 & 0.2 & 5 \\
\bottomrule
\end{tabular}
\caption{Hyperparameters used for model rollouts. All models were evaluated using VLLM with maj@k sampling, where k denotes the number of samples generated per problem. For expected success rate math datasets the Qwen series models use K=50.}
\label{tab:model_settings}
\end{table}

\section{Additional Routing Results and Data}
\label{app:routing_strategy_results}

\begin{table}[H]
\centering
\caption{Fireworks AI Pricing for routed model costs. Note GPT-OSS-20B has more than 16B params but uses its own pricing structure.}
\label{tab:fireworks_pricing}
\begin{tabular}{lc}
\hline
\textbf{Model Size / Type} & \textbf{Price (USD per million tokens)} \\
\hline
Less than 4B parameters & \$0.10 \\
4B -- 16B parameters & \$0.20 \\
More than 16B parameters & \$0.90 \\
OpenAI gpt-oss-20b & \$0.07 (input), \$0.30 (output) \\
\hline
\end{tabular}
\end{table}

\begin{table}[H]
\centering
\caption{Routing strategies comparison on MATH.}
\label{tab:MATH_routing_results}
\begin{tabular}{llrr}
\toprule
Strategy & Model & Accuracy & Cost \\
\midrule
\multicolumn{4}{l}{\textit{Single-model baselines}} \\
Math-7B & Qwen2.5-Math-7B-Instruct & 0.827 & 3.28 \\
gpt-oss-20b-low & gpt-oss-20b (low) & 0.866 & 4.00 \\
gpt-oss-20b-medium & gpt-oss-20b (medium) & 0.914 & 11.78 \\
R1-Qwen-7B & DeepSeek-R1-Distill-Qwen-7B & 0.914 & 18.43 \\
gpt-oss-20b-high & gpt-oss-20b (high) & 0.920 & 40.00 \\
\midrule
\multicolumn{4}{l}{\textit{Routing strategies}} \\
Random Routing & ensemble & 0.889 & 15.43 \\
Oracle (Perfect Knowledge) & ensemble & 0.957 & 6.37 \\
\midrule
\multicolumn{4}{l}{\textit{Probe router}} \\
Probe Router ($\lambda=0.00$) & ensemble & 0.930 & 28.37 \\
Probe Router ($\lambda=0.20$) & ensemble & 0.917 & 10.35 \\
Probe Router ($\lambda=0.40$) & ensemble & 0.909 & 8.29 \\
Probe Router ($\lambda=0.60$) & ensemble & 0.902 & 7.02 \\
Probe Router ($\lambda=0.80$) & ensemble & 0.894 & 5.98 \\
Probe Router ($\lambda=1.00$) & ensemble & 0.890 & 5.34 \\
\bottomrule
\end{tabular}
\end{table}

\begin{table}[H]
\centering
\caption{Routing strategies comparison on AIME 25.}
\label{tab:AIME 25_routing_results}
\begin{tabular}{llrr}
\toprule
Strategy & Model & Accuracy & Cost \\
\midrule
\multicolumn{4}{l}{\textit{Single-model baselines}} \\
Math-7B & Qwen2.5-Math-7B-Instruct & 0.100 & 0.04 \\
gpt-oss-20b-low & gpt-oss-20b (low) & 0.400 & 0.07 \\
R1-Qwen-7B & DeepSeek-R1-Distill-Qwen-7B & 0.567 & 0.43 \\
gpt-oss-20b-medium & gpt-oss-20b (medium) & 0.833 & 0.49 \\
gpt-oss-20b-high & gpt-oss-20b (high) & 0.933 & 1.74 \\
\midrule
\multicolumn{4}{l}{\textit{Routing strategies}} \\
Random Routing & ensemble & 0.433 & 0.52 \\
Oracle (Perfect Knowledge) & ensemble & 0.933 & 0.55 \\
\midrule
\multicolumn{4}{l}{\textit{Probe router}} \\
Probe Router ($\lambda=0.00$) & ensemble & 0.933 & 1.57 \\
Probe Router ($\lambda=0.20$) & ensemble & 0.867 & 0.74 \\
Probe Router ($\lambda=0.40$) & ensemble & 0.800 & 0.50 \\
Probe Router ($\lambda=0.60$) & ensemble & 0.733 & 0.44 \\
Probe Router ($\lambda=0.80$) & ensemble & 0.733 & 0.43 \\
Probe Router ($\lambda=1.00$) & ensemble & 0.700 & 0.39 \\
\bottomrule
\end{tabular}
\end{table}

\begin{table}[H]
\centering
\caption{Routing strategies comparison on GSM8K.}
\label{tab:GSM8K_routing_results}
\begin{tabular}{llrr}
\toprule
Strategy & Model & Accuracy & Cost \\
\midrule
\multicolumn{4}{l}{\textit{Single-model baselines}} \\
gpt-oss-20b-low & gpt-oss-20b (low) & 0.900 & 0.39 \\
Math-7B & Qwen2.5-Math-7B-Instruct & 0.945 & 0.43 \\
gpt-oss-20b-medium & gpt-oss-20b (medium) & 0.933 & 0.82 \\
R1-Qwen-7B & DeepSeek-R1-Distill-Qwen-7B & 0.884 & 1.39 \\
gpt-oss-20b-high & gpt-oss-20b (high) & 0.944 & 2.14 \\
\midrule
\multicolumn{4}{l}{\textit{Routing strategies}} \\
Random Routing & ensemble & 0.918 & 0.99 \\
Oracle (Perfect Knowledge) & ensemble & 0.978 & 0.40 \\
\midrule
\multicolumn{4}{l}{\textit{Probe router}} \\
Probe Router ($\lambda=0.00$) & ensemble & 0.940 & 0.95 \\
Probe Router ($\lambda=0.20$) & ensemble & 0.946 & 0.55 \\
Probe Router ($\lambda=0.40$) & ensemble & 0.945 & 0.47 \\
Probe Router ($\lambda=0.60$) & ensemble & 0.942 & 0.43 \\
Probe Router ($\lambda=0.80$) & ensemble & 0.941 & 0.42 \\
Probe Router ($\lambda=1.00$) & ensemble & 0.939 & 0.41 \\
\bottomrule
\end{tabular}
\end{table}

\subsection{Full-Cost Routing with an SC-Entropy Baseline}\label{app:full_cost_routing}
We revisit the routing results under more complete cost accounting. Under this formulation for probe-based routing, we include the input-token cost of obtaining activations from each candidate model, in addition to the selected model's generation cost. This captures the cost of computing the routing signal itself. Based on our observations, this cost is minimal, as we perform only one forward pass per model per question.

\paragraph{SC-Entropy Routing.}
This baseline replaces the probe-estimated success probability $\hat{p}_i(x)$
with a confidence score derived from self-consistency entropy (SC-Entropy),
while keeping the utility-based routing formulation of
Section~\ref{sec:utility_routing} unchanged. For model $M_i$, we compute the
Shannon entropy $H_i(x)$ over the answers extracted from its $K$ sampled
outputs and define

\[
c_{\mathrm{SC},i}(x)
=
1-\frac{H_i(x)}{\log K}.
\]
The selected model is therefore
\[
\hat{M}_{\mathrm{SC}}(x)
=
\arg\max_i
\left(
c_{\mathrm{SC},i}(x)-\lambda\hat{c}_i
\right).
\]

Unlike the probe score, this confidence signal is available only after all
$K$ outputs have been generated for every candidate model. We therefore
charge the SC-entropy baseline for the full generation cost across the model
pool. It serves as an expensive, output-based reference rather than a
practical cost-saving router.

\begin{table}[h]
\centering
\caption{GSM8K routing results. Model columns give the fraction of queries
routed to each model.}
\label{tab:routing-gsm8k}
\small
\setlength{\tabcolsep}{4pt}
\begin{tabular}{lrrrrrrr}
\toprule
\multirow{2}{*}{Strategy} & \multirow{2}{*}{Cost (\$)} & \multirow{2}{*}{Acc.}
& \multicolumn{5}{c}{Routing distribution} \\
\cmidrule(l){4-8}
& & & Math-7B & R1-Qwen-7B & oss-low & oss-med & oss-high \\
\midrule
Oracle                  & 0.4223 & 97.8\% & 6\%  & 1\%   & 92\% & 1\%  & 0\%    \\
Probe ($\lambda=1.00$)  & 0.4318 & 94.1\% & 39\% & 0\%   & 60\% & 1\%  & 0\%    \\
Probe ($\lambda=0.59$)  & 0.4588 & 94.4\% & 47\% & 0\%   & 51\% & 2\%  & 0\%    \\
Probe ($\lambda=0.53$)  & 0.4713 & 94.5\% & 47\% & 0\%   & 49\% & 3\%  & 0\%    \\
Probe ($\lambda=0.37$)  & 0.5170 & 94.5\% & 49\% & 0\%   & 43\% & 8\%  & 0\%    \\
Random                  & 0.9986 & 92.0\% & 20\% & 21\%  & 19\% & 21\% & 19\%   \\
SC-Entropy              & 5.2471 & 95.8\% & 94\% & 3\%   & 1\%  & 2\%  & 1\%    \\
MIRT Router             & 2.0130 & 94.1\% & 14.4\% & 1.5\% & 0.3\% & 6\% & 77.8\% \\
\bottomrule
\end{tabular}
\end{table}
 
%------------------------------------------------------------------
% Table 7: AIME routing
%------------------------------------------------------------------
\begin{table}[h]
\centering
\caption{AIME routing results. Model columns give the fraction of queries
routed to each model.}
\label{tab:routing-aime}
\small
\setlength{\tabcolsep}{4pt}
\begin{tabular}{lrrrrrrr}
\toprule
\multirow{2}{*}{Strategy} & \multirow{2}{*}{Cost (\$)} & \multirow{2}{*}{Acc.}
& \multicolumn{5}{c}{Routing distribution} \\
\cmidrule(l){4-8}
& & & Math-7B & R1-Qwen-7B & oss-low & oss-med & oss-high \\
\midrule
Oracle                  & 0.5490 & 93.3\% & 17\% & 17\% & 30\% & 27\% & 10\% \\
Random                  & 0.5187 & 43.3\% & 30\% & 30\% & 17\% & 10\% & 13\% \\
SC-Entropy (best)       & 2.7740 & 93.3\% & 0\%  & 30\% & 3\%  & 33\% & 33\% \\
Probe ($\lambda=0.98$)  & 0.4216 & 66.7\% & 3\%  & 10\% & 30\% & 57\% & 0\%  \\
Probe ($\lambda=0.27$)  & 0.5401 & 83.3\% & 0\%  & 13\% & 3\%  & 77\% & 7\%  \\
Probe ($\lambda=0.08$)  & 0.8607 & 86.7\% & 0\%  & 7\%  & 0\%  & 57\% & 37\% \\
Probe ($\lambda=0.04$)  & 1.1288 & 93.3\% & 0\%  & 3\%  & 0\%  & 43\% & 53\% \\
MIRT Router             & 0.9460 & 80.0\% & 0\%  & 0\%  & 3\%  & 70\% & 27\% \\
\bottomrule
\end{tabular}
\end{table}
 
%------------------------------------------------------------------
% Table 8: MATH routing
%------------------------------------------------------------------
\begin{table}[h]
\centering
\caption{MATH routing results. Model columns give the fraction of queries
routed to each model.}
\label{tab:routing-math}
\small
\setlength{\tabcolsep}{4pt}
\begin{tabular}{lrrrrrrr}
\toprule
\multirow{2}{*}{Strategy} & \multirow{2}{*}{Cost (\$)} & \multirow{2}{*}{Acc.}
& \multicolumn{5}{c}{Routing distribution} \\
\cmidrule(l){4-8}
& & & Math-7B & R1-Qwen-7B & oss-low & oss-med & oss-high \\
\midrule
Oracle                  & 6.4429  & 95.7\% & 87\% & 1\%  & 9\%  & 3\%  & 0\%  \\
Random                  & 15.7569 & 88.7\% & 20\% & 20\% & 20\% & 21\% & 20\% \\
SC-Entropy (best)       & 77.8804 & 92.4\% & 74\% & 20\% & 2\%  & 2\%  & 2\%  \\
Probe ($\lambda=1.00$)  & 5.5612  & 89.2\% & 61\% & 1\%  & 35\% & 3\%  & 0\%  \\
Probe ($\lambda=0.10$)  & 12.5633 & 92.1\% & 40\% & 16\% & 23\% & 21\% & 1\%  \\
Probe ($\lambda=0.02$)  & 20.4636 & 92.8\% & 28\% & 25\% & 12\% & 27\% & 9\%  \\
Probe ($\lambda=0.00$)  & 26.1820 & 93.1\% & 19\% & 26\% & 5\%  & 16\% & 33\% \\
MIRT Router             & 38.5990 & 92.0\% & 2\%  & 4\%  & 0\%  & 0\%  & 94\% \\
\bottomrule
\end{tabular}
\end{table}

\subsection{Cascade Routing}
\label{app:Cascade_Approach}

Let $M_s$ denote a base model and $M_l$ a stronger model with higher inference cost. 
For each input $x$, we use a threshold-based rule:
\[
M(x) = \begin{cases}
M_l & \text{if } \hat{p}_s(x) < \tau \\
M_s & \text{otherwise}
\end{cases}
\]
where $\hat{p}_s(x)$ is the probe's estimated probability that $M_s$ will answer correctly,
and $\tau \in [0,1]$ controls the tradeoff between performance and cost.
We evaluate using Qwen2.5-Math-1.5B as $M_s$ and Qwen2.5-Math-7B as $M_l$.

Our oracle in this setting iterates through models from cheapest to most expensive and routes to the cheapest model that solves the problem correctly, escalating only on actual failures. If no model succeeds, it defaults to the cheapest.

\begin{figure}[H]
    \centering
        \centering
        \includegraphics[width=0.5\linewidth]{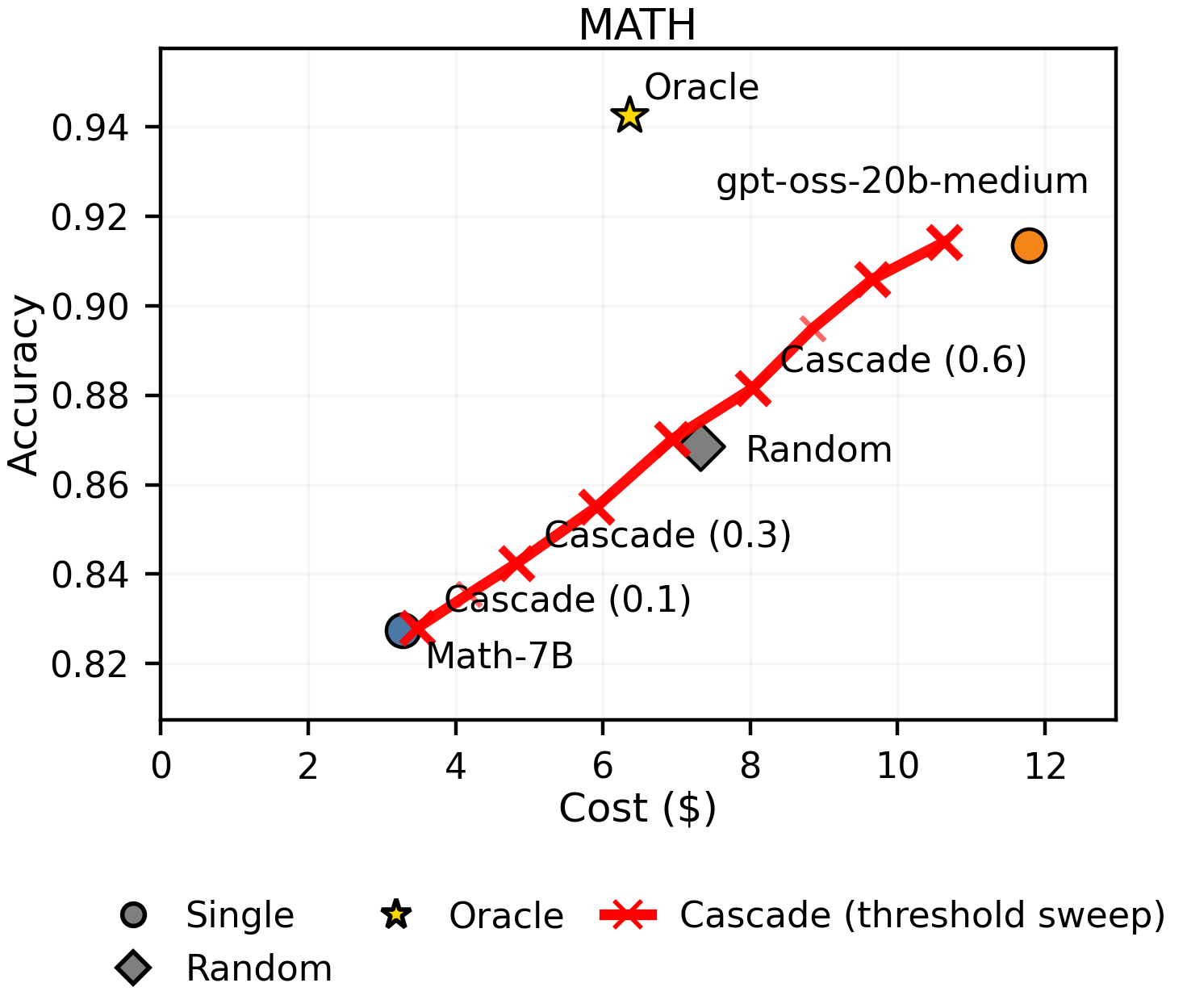}
        \caption{The cascade strategy (left) routes between Qwen2.5-Math-7B and GPT-OSS-20B-medium, substantially outperforming random allocation across the Pareto frontier. At $\tau{=}0.6$, the cascade matches GPT-OSS-20B-medium's 91.2\% accuracy while reducing cost by 17\%.}
\end{figure}

\end{document}